## Title

Full: Breaking Barriers in Robotic Soft Tissue Surgery: Conditional Autonomous Intestinal Anastomosis
Short: Conditional Autonomous Laparoscopic Surgery

## Authors

H. Saeidi [1,2], J. D. Opfermann [1,2], M. Kam [1,2], S. Wei [2,3], S. Leonard [2], M. H. Hsieh [4], J. U. Kang [2,3], A. Krieger [1,2]*

## Affiliations

[1] Department of Mechanical Engineering, Johns Hopkins University; Baltimore, MD 21211, USA.
[2] Laboratory for Computational Sensing and Robotics, Johns Hopkins University; Baltimore, MD 21211, USA.
[3] Department of Electrical and Computer Engineering, Johns Hopkins University; Baltimore, MD 21211, USA.
[4] Department of Urology, Children's National Hospital; 111 Michigan Ave. N.W., Washington, DC 20010, USA.
* Corresponding author: Axel Krieger, Email: axel@jhu.edu

## Abstract

Autonomous robotic surgery has the potential to provide efficacy, safety, and consistency independent of individual surgeon's skill and experience. Autonomous soft-tissue surgery in unstructured and deformable environments is especially challenging as it necessitates intricate imaging, tissue tracking and surgical planning techniques, as well as a precise execution via highly adaptable control strategies. In the laparoscopic setting, soft-tissue surgery is even more challenging due to the need for high maneuverability and repeatability under motion and vision constraints. We demonstrate the first robotic laparoscopic soft tissue surgery with a level of autonomy of 3 out of 5, which allows the operator to select among autonomously generated surgical plans while the robot executes a wide range of tasks independently. We also demonstrate the first *in vivo* autonomous robotic laparoscopic surgery via intestinal anastomosis on porcine models. We compared the criteria including needle placement corrections, suture spacing, suture bite size, completion time, lumen patency, and leak pressure between the developed system, manual laparoscopic surgery, and robot-assisted surgery (RAS). The ex vivo results indicate that our system outperforms expert surgeons and RAS techniques in terms of consistency and accuracy, and it leads to a remarkable anastomosis quality in living pigs. These results demonstrate that surgical robots exhibiting high levels of autonomy have the potential to improve consistency, patient outcomes, and access to a standard surgical technique.

## Summary

We demonstrate the first example of robotic soft tissue surgery with a level of autonomy of 3 under laparoscopic constraints.

**Main Text**

# INTRODUCTION

Autonomous robotic surgery systems have the potential to significantly improve efficiency, safety, and consistency over current tele-operated robotically assisted surgery (RAS) with systems such as the da Vinci robot (Intuitive Surgical Inc., Sunnyvale, CA, USA). Such systems aim to provide access to standard surgical solutions that are independent of individuals' experience and day-to-day performance changes. Proven examples of autonomous surgical robotic systems include TSolution-One *(1)* (THINK Surgical, Fremont, CA, USA) for procedures on rigid bony tissues, ARTAS for hair restoration *(2)* (Restoration Robotics Inc., San Jose, CA, USA), and Veebot *(3)* (Veebot LLC, Mountain View, CA, USA) for autonomous blood sampling. Currently, the most advanced autonomous capabilities are realized in the CyberKnife robot *(4)* (Accuracy Inc., Sunnyvale, CA, USA) which performs radiosurgery for brain and spine tumors under human supervision. However, this robot uses a contactless therapy method for tissues that are enclosed in rigid bony structures. Despite such efforts, autonomous soft tissue surgery still poses significant challenges.

Autonomous soft tissue surgery in unstructured environments requires accurate and reliable imaging systems for detecting and tracking the target tissue, complex task planning strategies that take tissue deformation into consideration, and precise execution of plans via dexterous robotic tools and control algorithms that are adaptable to the dynamic surgical situations *(5)*. In such highly variable environments, preoperative surgical planning such as in rigid tissues is not a viable solution *(6)*. In the case of laparoscopic surgeries, the difficulty further increases due to limited access and visibility of the target tissue and the disturbances from respiratory motion artifacts. Anastomosis is a soft tissue surgery task that involves the approximation and reconstruction of luminal structures and requires high maneuverability and repeatability and hence is a suitable candidate for examining autonomous robotic surgery systems in soft tissue surgery scenarios. Well over a million anastomoses are performed in the USA each year *(7–11)*. Critical factors affecting the anastomotic outcome include: the health of local tissue including perfusion status and contamination; physical parameters in anastomotic techniques such as suture bite size, spacing, and tension; anastomotic materials including suture and staples; and human factors such as the surgeon's technical decisions and experience *(12–17)*.

The level of autonomy (LoA) of medical robots, is categorized in different levels ranging from pure teleoperation to full autonomy *(5, 18)*. In this work, we utilize the classification introduced in *(19)*, including LoA 0: no autonomy (e.g., pure teleoperation), LoA 1: robot assistance (continuous control by human with some mechanical guidance or assistance from robot via virtual fixtures or active constraints *(20)*), LoA 2: task autonomy (robot autonomously performs specific tasks such as running sutures which are initiated by human via a discrete control rather than continuous control), LoA 3: Conditional Autonomy ("A system generates task strategies but relies on the human to select from among different strategies or to approve an autonomously selected strategy" *(19)*), LoA 4: High autonomy (robot makes medical decisions but it has to be supervised by a qualified doctor), and LoA 5: Full autonomy (a robotic surgeon that performs an entire surgery without the need of a human).

The most well-known example of LoA 0 autonomy is the da Vinci surgical system in which every motion of the robot during surgery is teleoperated by the surgeon via a master console. Many of the recent works such as SmartArm *(21, 22)* still include a LoA 0 autonomy and are developed for operation under space limitations such as neonatal chest surgery. LoA 1 is implemented in the form of shared control for reducing the complexity of steering flexible robotic endoscopes *(23)*. The first demonstration of LoA 2 (task autonomy) via *in vivo* open surgeries *(24)* was enabled via a robotic suturing tool controlled by a robot arm and a dual channel near infrared (NIR) and plenoptic 3D camera that allowed the robot to detect the target tissue (stabilized outside of the body) and its landmarks, calculate a linear suture plan on the tissue, and execute the suture placement step by step under human supervision. Recent methods demonstrate LoA 2 laparoscopic *in vivo* hernia repair for porcine models *(25)*. Machine learning-based techniques mostly automate surgical subtasks such as tumor ablation *(26)*, debridement of viscoelastic tissue phantoms *(27)*, clearing the surgical field *(28)* (i.e., LoA 2), or imitate surgical procedures via learning by observation on ex vivo phantom tissues in a LoA 3 *(29–32)*. Despite these considerable efforts, most works either present low autonomy in complex tasks or high autonomy in simpler task and phantom tissues. There is still a lack of methods enabling higher levels of autonomy (i.e., LoA 3 and LoA 4) in complex surgical tasks such as laparoscopic bowel anastomosis, while including proper human supervision to guarantee the success of surgery.

As our first contribution, we aim to achieve a LoA 3 based on the definition in *(19)*, i.e., conditional autonomy for robotic laparoscopic anastomosis. The technology for a LoA 3 requires several autonomous features including, starting/pausing/unpausing the tissue tracking system, detecting the breathing motion of tissue and its deformations and notifying the operator to initiate a replanning step, robot tool failure detection, camera motion control, suture planning in different modes with uniform and non-uniform spacing, pre-filtering the plan for reducing noise and irregularity, predicting the tool collisions with the tissue, and synchronizing the robot tool with breathing motions of the tissue and under a remote center of motion (RCM). The operator selects among the autonomously suggested suture plans or approves a re-planning step and monitors the robot for repeating a stitch as needed. Therefore, the LoA 3 developments provide a significantly higher autonomy during the surgical procedure compared to a step-by-step method for LoA 2.

The second contribution of this work is enabling a laparoscopic implementation of the conditional autonomy for full anastomosis, which imposes various technical and workflow challenges. In the new laparoscopic setting, the tissue is suspended in the peritoneal space of the patient with transabdominal stay sutures. In contrast to open surgical settings, where the motion of target tissue motion is restricted via external fixtures outside the body and the tissue is more accessible, the laparoscopic settings introduce multiple challenges including the effects of breathing motions on the staged tissue location and deformation, kinematic and spatial motion constrains for the robots through the laparoscopic ports, tool deflections and difficulties in force sensing, limited visibility, and a need for miniaturized laparoscopic 3D imaging systems, and other factors such as humidity during the surgery. We develop machine learning, computer vision, and advanced control techniques to track the target tissue while the patient is breathing, detect the tissue deformations between different suturing steps, and operate the robot under motion constraints. We demonstrate the first implementation of such soft tissue robotic surgeries in a series of *in vivo* survival studies via porcine models.

## RESULTS

In this section, we explain the design and workflow of the LoA 3 smart tissue autonomous robot (STAR), then present accuracy testing of the tissue motion tracker, comparison testing between STAR and expert surgeons performing complete end-to-end anastomosis in phantom bowel tissues, porcine in-vivo survival studies using STAR to complete laparoscopic small bowel anastomosis, and a marker less tissue landmark tracking method.

### System design for level of 3 autonomy

The laparoscopic version of STAR with a LoA 3 architecture is shown in Fig. 1. This system consists of one KUKA LBR Med robot with a motorized Endo 360 suturing tool *(33)* for robotic suturing and a second KUKA LBR Med robot that carries an endoscopic dual-camera system consisting of a NIR camera and a 3D mono color endoscope *(34)*. This camera system allows the STAR to autonomously track biocompatible NIR markers *(35)* on the tissue and reconstruct the 3D surface of the tissue (used for suture planning). The 2D view of the camera is also used as visual feedback for the operator when supervising the autonomous control (Fig. 1A).

The following describes the general workflow of the STAR with LoA 3 (for further details, please refer to the Materials and Methods section). First, the operator initiates a planning sequence with STAR using the graphical user interface (GUI) shown in Fig. 1B. The tissue tracking algorithm (detailed and tested in "Tissue motion tracking" Section) detects when the patient is temporarily not breathing (i.e., target tissue has reached a stop position due to a pause in the breathing cycle). When the tissue is stationary, STAR generates two initial suture plans to connect adjacent biocompatible NIR markers placed on the corners of the tissue. The suture plans are then filtered for noise and STAR predicts tool-tissue collisions. Unusable suture plans are discarded, and a new set of plans are generated for filtering and collision predictions. Once the set of plans is usable, the operator can select one of two suture plans: 1) Uniform spacing across the entire sample, or 2) uniform spacing with additional stitches in the corners of the tissue to elicit leak free closure in difficult geometry. Once the operator selects and approves one of the plans, STAR will execute a suture placement routine by moving the tool to the location of the first planned suture, applying a suture to the tissue, waiting for the assistant to clear loose suture from the field, and tension the suture. After completion of a suture routine, STAR reimages the surgical field to determine the amount of tissue deformation. If STAR detects a change in tissue position greater than 3mm compared to the surgical plan was approved by the operator, it will notify the operator to initiate a new suture planning and approval step. Otherwise, STAR will suggest reusing the initial surgical plan and move to the next suture routine. This process is continued for every suture in the surgical plan. In the instance where a planned stitch is not placed correctly or if the suture does not extend across two tissue layers, the operator can repeat the last stitch with a slight position adjustment via the GUI before the STAR continues to execute the suture plan.

**Tissue motion tracking**

In order to track the motion of the tissue due to breathing and other tissue motion during laparoscopic surgery, we developed a machine learning algorithm based on the convolutional neural networks (CNN) *(36)*and the feedback from the NIR camera. We labeled a total of 9294 examples of different motion profiles (4317 for the breathing examples and 4977 for the stopped breathing examples) via the data collected from a set of acute studies. The inputs to the CNN include the position history of the NIR markers in the past 2 seconds, and the direction of motion of the markers between now and 2 seconds ago which include 2 channels of 128 by 128 down sampled frames (see examples in Fig. 2A). The CNN includes 4 convolutional, 3 dense layers, and two outputs (for moving/stopped tissue labeling) which can predict the motion profiles with an accuracy of 93.56% during the training phase. The result is used to synchronize the motion of robot with tissue and trigger data collection and planning the right times (e.g., on the bottom of breathing cycles) to improve the control algorithm accuracy.

The results of the accuracy tests, and comparison with an optical flow method are shown in Fig. 2D. The optical flow methods track the differences of the last two frames in the NIR image, apply a low pass filter, and use a threshold to detect when the breathing motion ends (i.e., the image flow drops below a certain threshold). Since this method is sensitive to light intensity and distance, two variations are tested. A variation with a fixed threshold for a mid-range distance and light intensity (OF), as well as a variation with adjustable threshold for each measurement distance, (OA). For each test, we evaluated whether transitions between the breathing/moving state of the tissue and non-breathing state (i.e., at T1 and T2 in Fig. 2B) are detected correctly in a 14 breaths per minute breathing cycle. If the tested methods labeled any of the transitions incorrectly, a mistake was counted. Combinations of distance to the target tissue (3cm, 6.5cm and 10cm) for 11 marker orientations (Fig. 2C) were recorded for five breathing cycles for each combination using synthetic colon samples ($N = 165$ per condition). The results shown in Fig. 2D illustrate that the CNN-based method is more robust to tests in different conditions owing to the generalization capability of the neural networks. The CNN method results in 91.52% test accuracy which is significantly higher than OF with 75.76% ($P = 0.0068$) and OA with 48.48% accuracies ($P < 0.0001$). Accuracy of the CNN remained unchanged when *in vivo* anastomosis (N = 60 representative samples) was performed in four porcine models (91.52% for ex vivo vs 90.00% for *in vivo*) while the OA method's accuracy dropped to 58.33% from the 75.76% in the ex vivo tests due to a lack of robustness to variations in the test conditions.

**Ex vivo end-to-end anastomosis**

Ex vivo experiments for end-to-end anastomosis were conducted using phantom bowel tissues (by 3D Med, Franklin, OH, USA). The study groups included, conditional autonomous robotic anastomosis via STAR (n = 5), a manual laparoscopy (LAP, n = 4) method, and a da Vinci SI based robotic assisted teleoperation method (RAS, n = 4) (see Fig. S1 for the testbeds). For STAR tests, tissue orientation was set between -20 and 20 degrees with 10-degree increments for each sample, while LAP and RAS tissue orientations were randomly selected between -20 and 20 degrees. For all study groups, the test tissues were attached to a linear stage that was programmed to move with a 14 breath/minute frequency and 3mm amplitude to replicate the *in vivo* breathing cycles. For the STAR study group, a NIR marker tracking algorithm was used to trigger robot motion to the planned suture point, and to enforce that sutures are applied during the rest phase of the breathing cycle. A target suture spacing of 3 mm was selected based on our previous results in *(24)*, which considers the tissue thickness T, bite depth H, and suture spacing S for a leak-free anastomosis (see Fig. 1G in this reference). For the LAP and RAS study groups, surgeons were instructed to suture with the same spacing and consistency that they would perform in a human patient. To simulate tissue movement that occurs during *in vivo* surgeries, random deformation of the tissue was induced once during back wall suturing, and twice during front wall suturing by loosening and tensioning the stay sutures in a different direction for all study groups. LAP and RAS data was collected from four different surgeons from either George Washington University Hospital, D. C., USA or Children's National Hospital, D.C., USA. The order of surgeons was randomized with half of the participants starting with LAP followed by RAS tests and vice versa.

Individual STAR, LAP, and RAS tests were recorded and analyzed to measure the total time to complete knots (STAR, LAP, RAS), complete sutures (STAR, LAP, RAS), planning (STAR), and supervision (STAR). Suture planning was defined as moving the camera to measure distance, calculating and saving suture plans, and returning back to the safe suture distance. Supervision was defined as slight position adjustments for the robot after the hesitancy events. The number of suture hesitancy events per stitch was defined as needle placement corrections for STAR or when surgeons inserted and immediately retracted a needle from tissue for LAP and RAS experiments. Furthermore, suture spacing, defined as the shortest distance between any two consecutive sutures, and bite depth, defined as the perpendicular distance from a suture to the tissue edge, were collected for each sample (see Fig. S2A). We report the coefficient of variance for suture spacing and bite depth as a measure of consistency for each metric since both are known to directly affect the quality of a leak-free anastomosis *(37)*. All results from ex vivo testing are summarized in Fig. 3.

STAR completed a total of 118 stitches (10 knots, 108 running stitches, N=5) with an average suture hesitancy per stitch of 0.17 ± 0.44. The total number of suture hesitancy events among all STAR samples was 20 indicating that STAR correctly placed stitches under the conditional autonomous mode at the first attempt 83.05% of the time. Twelve of the hesitancy events occurred during a corner stich, while the remaining eight hesitancy events occurred on either the back or front wall of the tissue. Observed hesitancy events were a result from small planning errors, tool positioning errors, difficult to reach corner stitches, and shallow bite depth. In each of these cases, the operator would interrupt the suture plan to correct the positioning of the suture tool, and then restart the conditional autonomous mode. The average three-dimensional norm of offset adjustments was 3.1 ± 1.39 mm with a maximum value of 6.0 mm. For the LAP method, a total of 111 stitches (18 knots, 93 running stitches, N=4) were completed with an average hesitancy per stitch of 1.03 ± 1.71 while RAS samples had a total of 139 stitches (14 knots, 125 running stitches,

N=4) with an average hesitancy per stitch of 0.44 ± 0.97. As summarized in Fig. 3A, the hesitancy per stitch for STAR was significantly less than that of LAP (P < 0.0001) and RAS (P = 0.0177).

The average total time to complete the end-to-end anastomosis task is shown in Fig. 3B. It was observed that RAS was the fastest method for completing the task (31.96 ± 9.01 minutes), followed by LAP (51.08 ± 32.72 minutes), and STAR (55.41 ± 2.94 minutes). While STAR was found to be similar to total task time to LAP (P = 0.8091), STAR was significantly slower than RAS (P=0.0009). Differences in total completion time are primarily due to additional re-planning due to random tissue deformations and motions, and overall supervision time (approving plans and sporadic slight position adjustments) in the STAR system that are not needed in the RAS technique which account for an additional 19.66 minutes per sample. Additionally, STAR is operated under conservative velocity limits to minimize the force interactions between the surgical tool and the laparoscopic ports to guarantee the smoothness of motion profiles and the accuracy of the suture positioning. The smooth motion profiles have the added benefit of reducing the chance that small planning errors in tight corners cause the suture tool to damage the stay sutures or native tissues.

The average suture spacing for LAP, RAS and STAR was 2.28 ± 1.04 mm, 1.58 ± 0.65 mm, and 3.05 ± 0.8 mm, respectively (see Fig. 3C). Suture spacing resulting from STAR was significantly larger than LAP (P < 0.0001) and RAS (P < 0.0001) but was statistically similar to the 3.0 mm spacing suggested by the uniform suture plan (P = 0.4163). Because smaller suture spacing can lead to smaller variance in the data, suture consistency is compared by the coefficient of variance for each study group (i.e., the ratio of standard deviation to mean converted to percentage). Based on the results in Fig. 3D, the coefficient of variance for LAP, RAS, and STAR was 45.37%, 41.42%, and 26.36% respectively, which indicates higher consistency of STAR compared to LAP (P<0.0001) and RAS (P<0.0001). The average bite depth for LAP, RAS, and STAR was 2.02 ± 1.10 mm, 1.69 ± 0.55 mm, and 3.05 ± 0.91 (see Fig. 3E). While the average bite depth for STAR was significantly larger than LAP (P < 0.0001), and RAS (P < 0.0001), it was similar to the target bite depth of 3 mm programmed by the uniform suture plan (P = 0.4414). As with suture spacing, the consistency of the bite depth can be reported as the coefficient of variation by normalizing the standard deviation of each study group by its mean. As illustrated in Fig. 3F, the coefficient of variation for LAP, RAS, and STAR was 54.41%, 32.57%, and 29.99% respectively, indicating that STAR is more consistent in bite depth than LAP (P < 0.0001) and similar to RAS (P = 0.2611) (see Table S1 for further details).

A representative anastomosis for each study group is shown in Fig. 3G. Back wall suturing is illustrated in the top row, front wall suturing is illustrated in the middle row, and three-dimensional flow fields of fluid passing through the resulting anastomosis are illustrated in the third row. To obtain images of the resulting flow fields, samples from each study group were connected to a MR conditional pump and circulated with a 60/40 glycerin to water mixture at a flow rate of 10 ml/s. Flow fields were acquired using a Siemens Aera 1.5T MRI scanner programmed to a 4D flow sequence with 1.25 mm x 1.25 mm x 1.3 mm spatial resolution, with the samples placed at the isocenter of the magnet. Post processing, including segmentation and display of flow field lines was performed with iTFlow version 1.9.40 (Cardio Flow Design Inc, Tokyo, Japan). A single phase of the 4D flow is presented for all models. Resulting flow lines are most laminar for STAR and most turbulent for LAP, indicating that STAR reapproximates tissue better than either surgical technique.

### *In vivo* end-to-end anastomosis

Finally, laparoscopic autonomous surgery was performed *in vivo* on porcine small bowel using STAR (n = 4). After surgery, the animals were monitored for a one-week survival period, and subject to limited necropsy to compare the healed anastomosis to a laparoscopic control (n = 1) (Table 1). For these tests, STAR utilized the same suture algorithm described in the ex vivo tests. In order to guarantee the safety of the procedure and comply with animal care protocols, we increased the safety measures by adding to the number of tasks that the operator is required to monitor and intervene (the ex-vivo tests already allowed higher risk tolerance for testing the system features). This included additional number of human interventions for initial robot position fine-tuning (before the first suture) and approval of firing the needle. We also reduced the number of plan options and replanning steps and implemented more conservative velocity limits for further preventing collateral tissue damage.

Upon completion of the survival animal, limited necropsy was performed. Gross examination of the anastomosis identified scar tissue and adhesions consistent with the control animal. Additionally, residual NIR markers used during the surgery had to be encapsulated and remained outside of the anastomosis. A portion of the small bowel was then resected and subject to leak and lumen patency testing. Lumen patency was defined as the largest diameter gauge pin that could be passed through the anastomosis. Leak pressure was defined as the maximum internal pressure achieved by the anastomosis before rupture. Leak pressure testing was limited to a maximum of 1.2 psi to preserve the sample for histology testing (see Fig. 4).

In this study, STAR achieved a leak pressure of 0.69 ± 0.59 psi and a lumen patency of 88.75 ± 4.79 percent which are both consistent with our previous results in open surgeries *(24)*. The average task completion time for STAR was 62.03 ± 5.32 minutes compared to 25.6 minutes for the control. Similar to the ex vivo tests, STAR was slower to complete the task primarily due to the additional planning (8.5 minutes on average for 3 plans), safe tool rotations (1.13 minutes), and task supervisions (10.2 minutes). Moreover, STAR was operated under conservative speed limits to minimize the interaction forces at the laparoscopic ports and reduce the chance of collisions between the suture tool, stay sutures, and non-targeted tissue due to breathing motions which further increased the total time. Histology on day 7 (Fig. 4A) illustrates that there was no observable difference in wound healing and inflammatory changes as observed in continuity of mucosal edges, degree of tissue injury such as hematoma, and similar counts of polymorphic nucleated cell (PMN) infiltrates (P = 0.4543) (Fig. 4B) between STAR and the control.

In all four *in vivo* tests, STAR completed a total of 86 stitches (8 knots, 78 running stitches). The total number of sutures requiring manual adjustments were 29 (66.28% correctly placed stitches in the first attempt) which corresponds to an average of 0.34 suture hesitancy per stitch. These attempts included 14 corner stitches and 15 stitches along the back and front walls. Compared to the ex vivo tests, a higher average number of attempts were required to adjust the offsets manually via the GUI (0.34 additional attempts for *in vivo* vs 0.17 for ex vivo). Additional sources of extra attempts observed during the *in vivo* experiments included animal size variations that affect the optimal laparoscopic port placement and tissue reachability, extra tissue motion induced by the stay sutures on the flexible skin of animal, sporadic insufflation leaks due to the motion of robot tool as well as imperfect sealing of the gel port, edema causing the tissue not to fit in the jaw, harder tissue staging depending on the animal, and tool flex under forces from the trocar. The average norm of offset adjustments in three dimensions was 3.85 ± 1.88 mm with a maximum value of 8.15 mm. For the control test condition, a total of 21 stitches were completed (4 knots and 17 running stitches). The total number of needle repositioning was 9 (57.14% correctly placed stitches in the first attempt), which corresponds to an average of 0.43 suture hesitancy per stitch. These needle repositioning adjustments included 4 corner stitches and 5 stitches on the front and back walls.

**Cascaded U-Net for segmentation assisted landmark detection**

To minimize the complexity of the camera system and also to eliminate the need for markers which are difficult to get approved for long-term implantation in clinical translation, we aim to replace the fluorescence marker-based imaging with a convolution neural network (CNN)-based landmark detection algorithm. The overall data processing contains a CNN architecture and a post-processing step as shown in Fig. 5A. The CNN architecture is adopted from U-Net, since U-Net has demonstrated superior performances in the segmentation task due to its capability of the multi-resolution analysis *(36)*, and it requires much less training data compared to a fully connected network due to its fully convolutional properties *(38)*. Two U-Nets are cascaded for segmentation assisted landmark detection. The first U-Net takes the intestine image as the input and it outputs a segmentation result of the intestine tissue. Its main architecture is mostly the same as the one in *(36)*, except that we use the same padding in all pooling layers to avoid image cropping, and that we use sigmoid activation instead of the soft-max activation for the final segmentation results. The second U-Net takes both the intestine image and the segmentation result from the first U-Net as the input and it outputs a landmark heatmap, since the landmark heatmap regression has demonstrated better performances than regressing the landmark coordinates directly *(39)*. The final landmark coordinates are obtained as the maximum responses of the predicted heatmaps. The second U-Net architecture is mostly the same as the first U-Net, except that its input contains two channels. Since this cascaded U-Net has two outputs, the total loss function is the summation of the binary cross entropy of both the segmentation and the landmark heatmap with the same weight. In the post-processing step, we first binarize the segmentation results by applying a threshold. Then, we find the largest connected component (CC) from the binarized segmentation. In the end, we point-wise multiply the largest CC with the landmark heatmap to eliminate any random landmarks detected on the background.

To evaluate the performance of the algorithm, we record a 45-seconds video of an ex vivo intestine tissue with different positions and orientations using our camera system. The video has a frame rate of 68 Hz and contains 3037 frames in total. We take 50 image frames out from the video every 0.9 second. The 25 image frames with the odd serial number are used as the training set, while the remaining 25 frames with even serial number are used as the testing set. We manually segment the intestine tissue using a binarized hard mask as the segmentation ground truth. Similarly, we manually segment four corners of the intestine tissue using binarized hard masks, and we use the center of each corner segment as the ground truth for the corner landmark coordinate (i.e., for replacing the NIR markers used for path planning). Moreover, we fit each corner hard mask with an elliptical gaussian probability as the heatmap ground truth. After training, the trained weights are applied to both the training set and the testing set followed by the post-processing step. The predicted landmark coordinates of the four corners are obtained as the four local maximum heatmap responses. Compared with the ground truth, it is found that the landmark detection accuracy to the training set is 2.3±1.0 pixels, while the landmark detection accuracy to the testing set is 2.6±2.1 pixels. Due to the different shapes and sizes of the four intestine corners, we define the maximum allowable error as the statistical average of the effective radius of all the corners, which is found to be 11.3±4.8 pixels. Furthermore, out of 25 testing frames, there are 3 frames where apart from the correct four corners, extra landmarks in the background are mistakenly detected. With the segmentation assisted landmark detection post-processing, these extra landmarks are effectively removed. The final segmentation and heatmap results are cropped and combined with the intestine image to a color image as shown in Fig. 5B. From these results, we conclude that the cascaded U-Net can correctly segment the intestine tissue and generate the intestine landmark heatmap. Moreover, with the segmentation assisted landmark detection post-processing, the noise from the background can be effectively removed.

## DISCUSSION

In this paper, we demonstrated the first conditional autonomous soft tissue surgery in unstructured and deformable environments under motion and visual constraints. Advanced imaging systems, machine vision and machine learning techniques, and real-time control strategies were developed and tested to track tissue position and deformation, perform complex surgical planning, interact with the human user, and adaptively execute the surgical plans. In the robotic laparoscopic anastomosis experiments, the developed system outperformed surgeons using LAP and RAS surgical techniques in metrics including the consistency of suture spacing and bite depth as well as the number of suture hesitancy events which directly affect the quality of a leak-free end-to-end anastomosis. The achieved LoA 3 encapsulates the autonomous control functionality and reduces the involvement of the human operator compared to the LoA 2 (task autonomy level) with the robot reacting to tissue deformation and various test condition changes. Although the role of human supervision cannot be eliminated in the complex and unpredictable surgical scenarios, advancing through the levels of autonomy can reduce the individual-based experience and performance variations and result in higher safety, access to a standard surgical technique, and consistency of the surgical outcomes. We also demonstrated the first *in vivo* robotic laparoscopic anastomosis which involved surmounting several challenges including soft tissue tracking, surgical planning, and execution in a highly dynamic and variable environment with restricted access and visibility. The survival study results indicated that the developed robotic system could match the performance of expert surgeons in metrics including leak-free anastomosis and lumen patency and at the same time exhibit a high level of consistency.

In the current version of the system, the laparoscopic dual-camera system has a 3 cm footprint which requires the use of a gel port for camera insertions. This will also increase the chance of gel port blocking the robotic suture tool or inducing extra pressure and slight positioning errors at the tool control point (TCP). In our future works, we will integrate and test the marker-less tissue tracking techniques in order to reduce the camera system to a 3D endoscope with a smaller footprint. Furthermore, we will enable a larger distance for accurate point cloud collection so that the camera system does not require back and forth motion between the imaging and suture positions to reduce the suturing time. It is also worth mentioning that due to additional forces and torques introduced at the laparoscopic port on the robot tool, the force measurements at the tool tip are not observable. More specifically, the 6-axis force/torque sensor of the robot is mounted on the robot flange which measures a combination of tool gravity forces (3 estimated forces based on the tool geometry and mass), interactions at the laparoscopic port (2 unknown forces and 2 unknown torques assuming negligible axial force and torque along the port), and the tension and contact forces at the tool-tip during the suturing process on tissue (3 unknown forces, i.e. a total of at least 7 unknowns with 6 measurements). In the future works, will include a tactile or proximity sensor at near the tool tip to decouple the interaction forces and measure the tissue contact and suture tensioning forces locally. Furthermore, we will add a sensor at the tool tip to detect if the two layers of the target tissue are inside the robot tool jaw before firing the needle. This will help eliminate one monitoring task from the operators' tasks and guarantee that robot does not miss any stitches on the tissue. We expect that the synergy of these changes will allow a significantly faster task completion time via higher (but safe) robot velocity limits, reduced measurement and (re)planning time, and reduced number of suture hesitancy events.

## MATERIALS AND METHODS

### Imaging System

A customized laparoscopic imaging system is built for in vivo animal studies via STAR (shown in Fig. 1A). The projector and the monochrome camera are used to reconstruct the 3D point cloud of the sample. The NIR light source and the NIR camera are used for fluorescence marker imaging. Both cameras are enabled simultaneously through the nonoverlapped imaging spectral window. Moreover, a white light source is added in order to monitor the environment inside the animal body when necessary. NIR and 3D cameras enable the robot to reconstruct the 3D model of the tissue and plan for robotic suture planning. The detailed specifications for the imaging system are presented in the supplementary methods.

### Robotic Platform

The platform consists of a KUKA LBR Med lightweight robot that is equipped with a motorized commercial Endo 360 suture tool with pitch control. The camera system allows measurement of the tissue geometry in a 5-8 cm distance from the tissue. The system is capable of dynamic position control of the camera system via a linear stage, which is used to prevent collisions between the camera system and suturing tool when STAR is executing a suture plan. In practice, the camera and suture tool are always in one of two modes depending on the status of the surgical plan: i) Imaging mode: The camera is positioned 5-8cm from the target tissue and used to collect 3D point clouds, while the suture tool is retracted from the field of view to prevent occlusion and collision with the surgical field. ii) Suture mode: The camera is positioned at a 4 cm distance behind the measurement position so the suture tool can be advanced into the surgical field and execute the surgical plan without collision with the imaging system.

**High-level Control Strategy**

In this control loop, real-time video frames from the 3D endoscope and NIR cameras are collected and processed via a raytracing technique that were developed in *(35)*. A tissue motion tracking algorithm (detailed and tested in the Results Section) tracks the position of target tissue via the NIR fluorescent markers in the NIR view of the camera, and autonomously detects when the tissue is stationary. At this point, the motion tracker triggers the imaging system to reconstruct the surgical field based on the image of the stationary scene. A path planning algorithm, developed in our previous works *(40, 41)*, generates multiple suture plan options on the 3D point cloud of the tissue which can be selected by the user from a graphical user interface (GUI). This method also projects the robot tool on each planned suture point and predicts the chance of tool collision with the tissue, and autonomously generates a new suture plan if the original image is noisy. Once STAR generates a usable suture plan that is selected by the operator, reference waypoints and real-time robot positions are used in a high-level suture logic and task planner via the methods developed in *(42)*. This planner generates the strategy for knots, running stitch indices, and also produces motion primitives (such as approaching the tissue, biting/firing the suture, waiting for the assistant, and tensioning).

**Low-level Control**

The motion primitives are then sent to a trajectory generator to obtain smooth time-based trajectories via Reflexxes Motion Libraries *(43)*. Kinematics and Dynamics Library (KDL) of Open Robot Control Software (OROCOS) convert the task-space trajectories of the robot into to the joint-space trajectories *(44)*. A remote center of motion (RCM) constraint is also implemented in the kinematic solver at the site of the laparoscopic tool port. Finally, low-level closed-loop controllers enabled via KUKA robot controllers in IIWA stack *(45)*, OROCOS Real-time toolkit (RTT) *(46)*, and Fast Research Interface (FRI) *(47)*, guarantee the joint-space trajectories control of the robots and the mounted suture tool and linear stage.

**Suture planning**

The suture planning and tissue deformation tracking logic is shown in Fig. 6. This method allows interactions with the operator for approving a suture plan or initiating a (re)planning step. The workflow is according to the following. The breathing motion tracker tracks the motion of tissue at the measure mode/distance. When an update plan command is issued by the operator, the motion tracker detects the end of breathing motion via a sequence of NIR image frames and triggers a 3D point cloud collection from the target tissue. Collecting the point cloud when the tissue is not moving is essential to remove blurry and noisy data that is caused by the motion of the tissue during the fringe projection process. The projected NIR marker positions on the point cloud (via raytracing) are used for determining the start and end points of the suturing path. For the backwall suturing, the NIR marker order is detected automatically via blob tracking techniques such that the top marker is identified first, followed by the left marker, and then the right marker. For the front wall, the order is left marker followed by right marker. The suture path planning method then plans suture points in between the NIR markers in the sequence mentioned above with 3 mm spacings and an optional extra corner stitch for preventing leaks at the corners. For each of the planned suture points, the system autonomously detects potential collisions of the suture tool with the target tissue by projecting the geometry of suture tool with a fixed bounding box onto the point cloud data from the camera. If the ratio between the number of points predicted to collide with the tool and the number of points projected to correctly fit inside the tool jaw exceeds an 80% threshold, the system warns the operator that suture misplacement or tool collision is possible, and the operator can select to generate a new suture plan.

**Tissue Deformation and Motion Tracking**

Once a suture plan is selected by the operator, the robot saves the plan and autonomously switches to the suture mode. At the end of the next breathing cycle, a first snapshot of the NIR view is recorded and the suturing process starts. Once a suture has been placed by STAR, the system waits until the end of the next breathing cycle and the camera system captures a new image via the NIR camera. The new image is then compared to the initial image, and if the norm distance of any marker exceeds 3 mm, STAR suggests to the operator that a new suture plan should be generated. A new suture plan is recommended in this instance since a tissue deformation greater than 3 mm exceeds more than half of the tool jaw resulting in suture positioning errors if the previous plan is used. If the detected motion between images is less than 3mm, a message is shown to the operator indicating that the previous plan is still usable, and the operator can continue to use the existing plan. Finally, the motion of the robot is synchronized with the breathing motion such that suture tool reaches the target tissue at the stationary point of the breathing cycle. The synchronization is achieved by calculating a trigger based on the current time $t$, distance to target $d$, the average robot velocity $v$ and the breathing cycle duration $T$. For the robot to reach the target at the $n$th breathing cycle, the trigger time can be calculated according to $t_t = nT - \frac{d}{v}$, where $n$ is $\min_n(nT - \frac{d}{v}) > t$.

**Surgical methodology**

The anesthetized animal was positioned supine, and the abdominal/surgical site was prepped with alcohol, followed by chlorhexidine scrub and sterile drapes. On the abdomen, four trocars (Ethicon, Somerville, NJ) were placed laterally, and a gel port (Applied Medical, Rancho Santa Margarita, CA) was inserted medially (e.g., as shown in Fig. S2B). The trocars provided sterile access to the peritoneal space for STAR's suture tool and the assistant's grasper, while the gel port provided sterile entry for the laparoscopic camera. STAR's suture tool and imaging system were disinfected using MetriCide 28 (Metrex, Orange, CA). Using insufflation and laparoscopic technique, a surgeon identified and transected a loop of small bowel. The two open ends of the intestine were reapproximated and suspended with transabdominal stay sutures (e.g., as shown in Fig. S2C). Near Infrared (NIR) markers (indocyanine green and Permabond) that can be visualized by STAR's laparoscopic camera were placed on the corners of the tissue, and robotic suturing was performed. After the anastomosis was completed, the abdominal wall was closed in a multi-layer fashion. The total procedure time was approximately 4 hours. See supplementary methods for the preoperative setup and the postoperative care.

**Statistical analysis**

GraphPad Prism 7.04 statistical software was used for all analysis in this study. Suture spacing and bite depth for each modality were subjected to the D'Agostino & Pearson test for normality and then compared using the non-parametric Mann-Whitney comparison test. Mann-Whitney tests are also used to compare the hesitancy per stitch, and time per stitch between STAR and the surgical modalities. Levene's test for variance was performed on total suture times, followed by Welch's t-test assuming unequal variance to compare STAR and LAP total suture times, and the unpaired t-test assuming equal variance to compare STAR and RAS total suture times. Single comparisons are chosen for all comparison tests because our hypothesis only considers relationships between STAR and each of the surgical techniques individually. Variation in suture spacing and bite depth for each test modality is normalized using the sample mean and reported as the coefficient of variation, with statistical differences calculated as in *(48)*. A Kruskal-Wallis test for analysis of variance followed by Dunnett's tests for multiple comparisons was performed for ex vivo motion tracking accuracy, while Mann-Whitney comparison tests were performed for in vivo tracking accuracy. Unpaired t-tests are used for comparing PMN counts. P-values are reported for all comparison tests, with $P < 0.05$ considered to be statistically significant.

**Acknowledgments:** The authors would like to thank Mr. Jiawei Ge and Vincent Cleveland for their help with material preparation and analyzing the flow data in iTFlow software. We would also like to thank Dr. Yun Chen, Mr. Matthew Pittman, and Junjie Chen for their help with digitizing histology slide. Finally, we would like to thank the surgeons from the George Washington University Hospital, D. C., USA and Children's National Hospital, D.C., USA, who participated in the study.

**Funding:** Research reported in this paper was supported by National Institute of Biomedical Imaging and Bioengineering of the National Institutes of Health under award numbers 1R01EB020610, and R21EB024707. The content is solely the responsibility of the authors and does not necessarily represent the official views of the National Institutes of Health.

**Author contributions:**
Conceptualization: JDO, SL, JUK, AK
Methodology: HS, JDO, MK, SW, SL, MHH, JUK, AK
Software: HS, MK, SW, SL
Investigation: HS, JDO, MK, SW, SL, MHH, JUK, AK
Visualization: HS, JDO, MK, SW, AK
Data Curation: HS, JDO, MK, SW
Formal Analysis: HS, JDO, MK, SW
Funding acquisition: MHH, JUK, AK
Supervision: MHH, JUK, AK
Writing – original draft: HS, JDO, MK, SW
Writing – review & editing: HS, JDO, MK, SW, SL, MHH, JUK, AK

**Competing interests:** Authors declare that they have no competing interests.

**Data and materials availability:** All data are available in the main text or the supplementary materials.

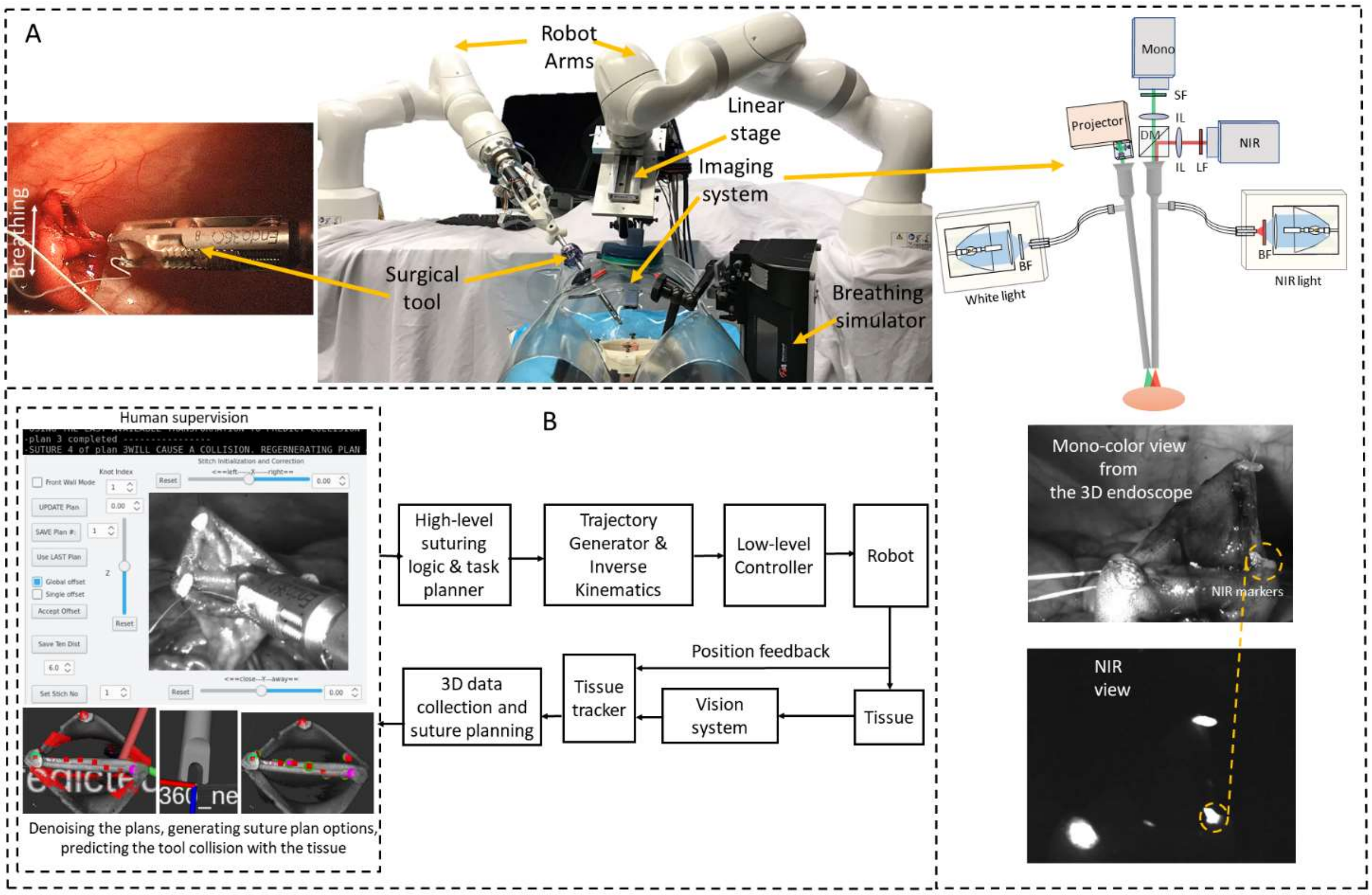

**Fig. 1. Conditional autonomous laparoscopic soft tissue surgery.** (A) The components of the STAR system including medical robotic arms, actuated surgical tools, and dual-channel NIR and 3D structured light endoscopic imaging system. (B) Control architecture of the conditional autonomous control strategy (level of autonomy 3) for STAR.

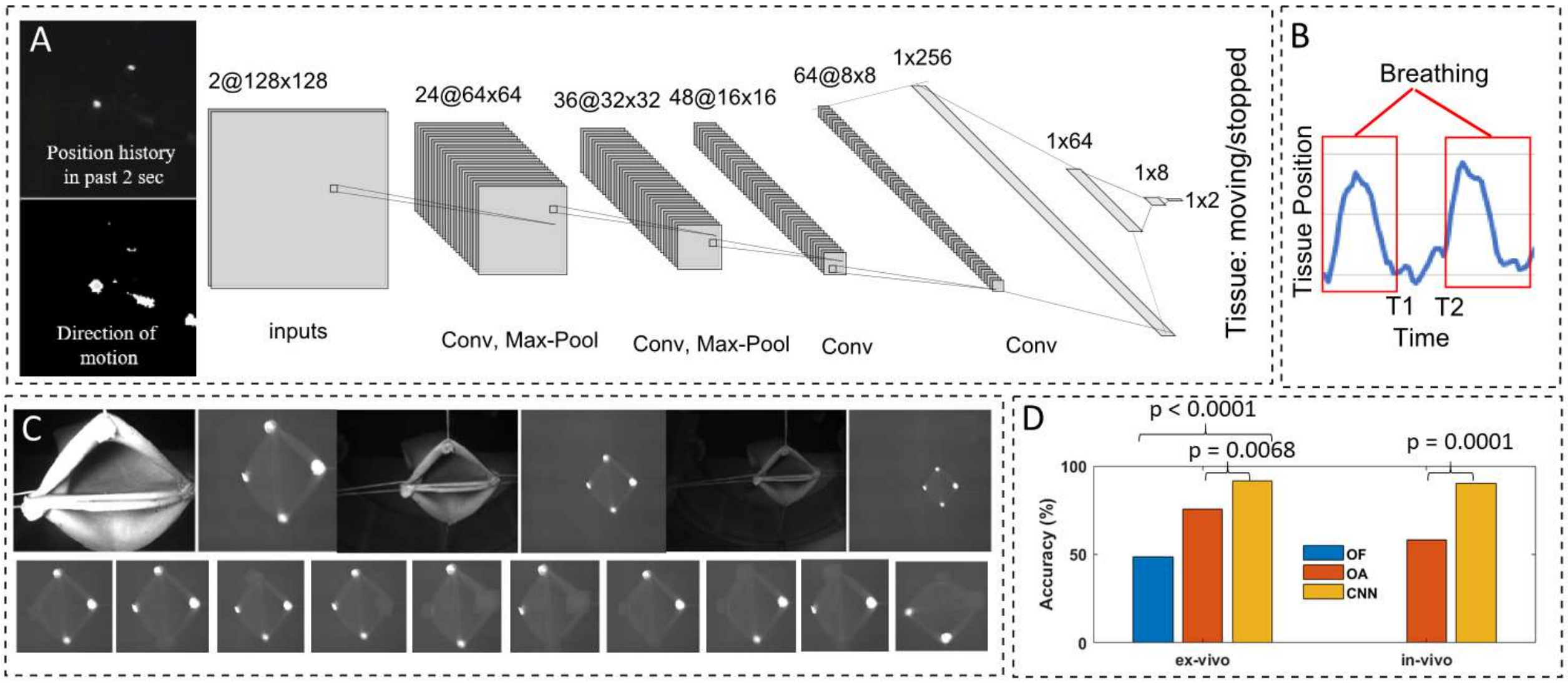

**Fig. 2. Tissue motion tracking.** (A) The CNN-based breathing motion tracker. (B) Examples of the vertical motion of NIR marker during *in vivo* tests. (C) Robustness test configurations for the ex vivo conditions. (D) The accuracy test results for the breathing motion tracker via optical flow with fixed threshold (OF), optical flow with adjustable threshold (OA), and the CNN-based method (CNN).

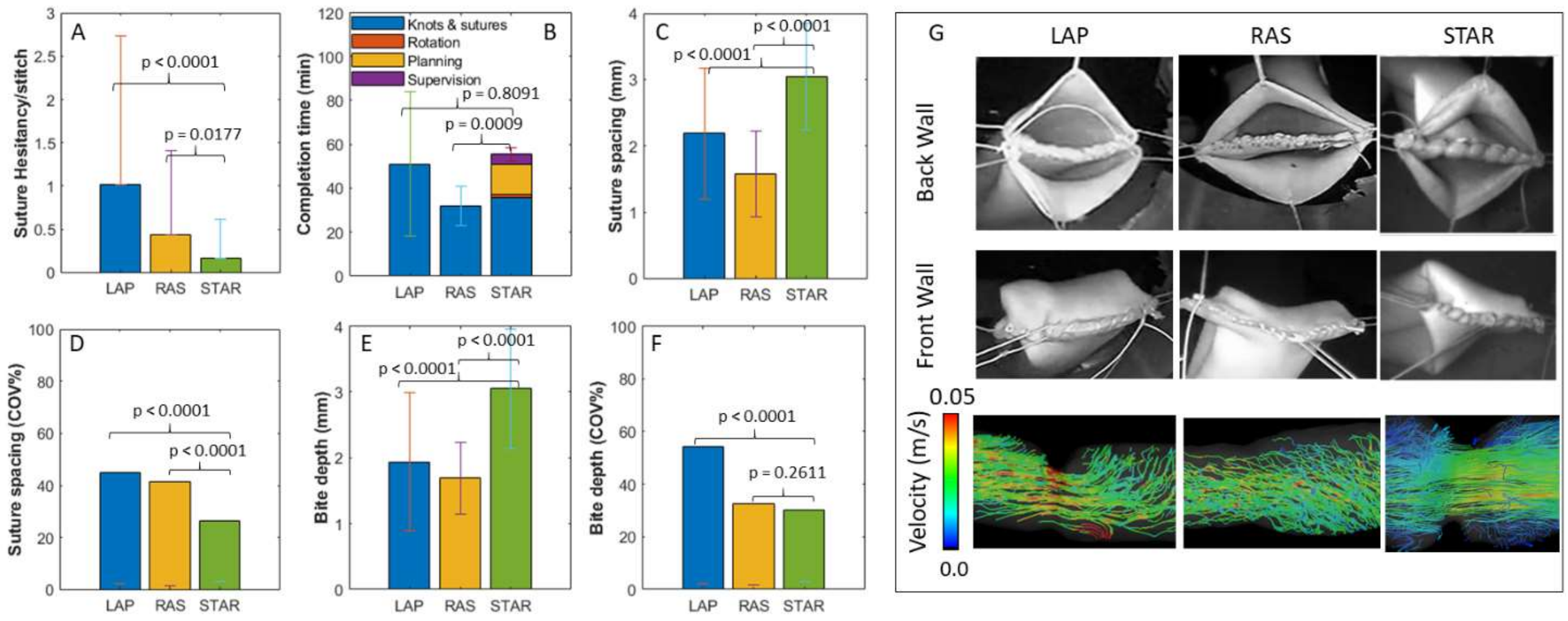

**Fig. 3. The results of ex vivo end-to-end anastomosis via LAP (n = 4), RAS (n = 4), and STAR (n = 5).** (A) Suture hesitancy events per stitch (additional suturing attempts per stitch). (B) Task completion time. (C) Suture spacing. (D) consistency of suture spacing via the coefficient of variance (COV). (E) Suture bite depth. (F) Consistency of bite depth via the coefficient of variance (COV). (G) Representative examples of the ex vivo end-to-end anastomosis test via LAP, RAS, and STAR including three-dimensional flow fields within each sample.

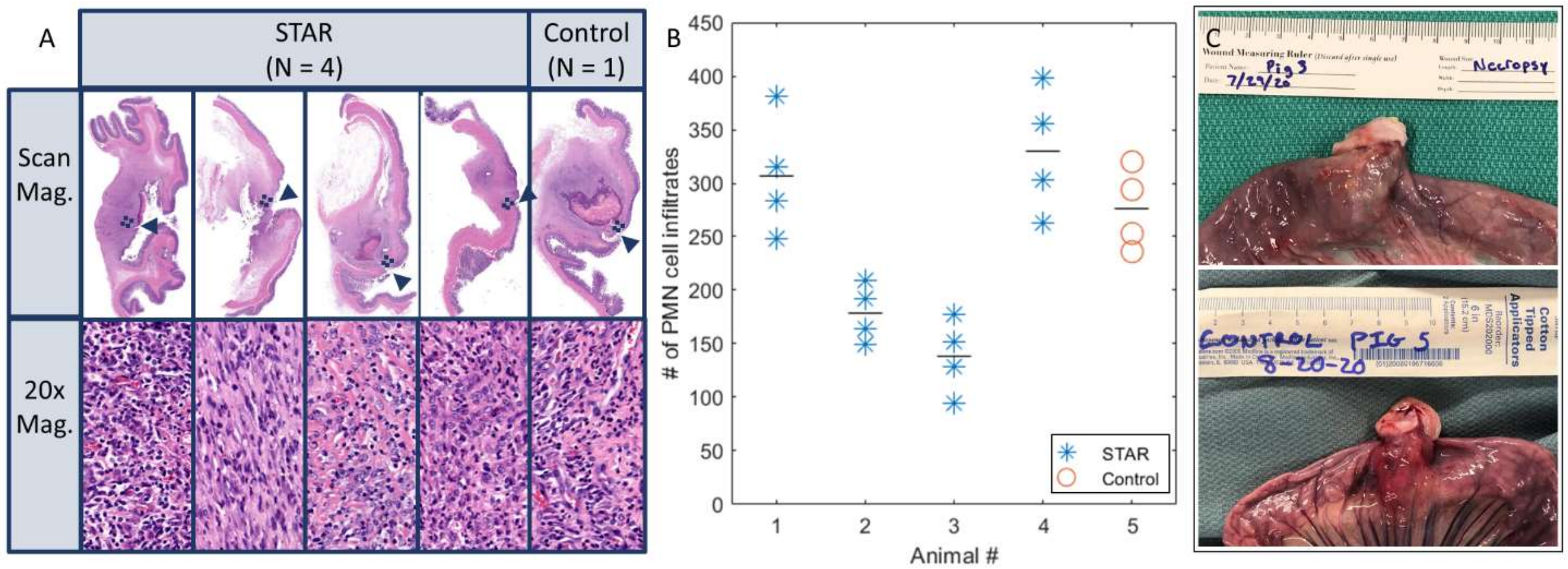

**Fig. 4. The results of *in vivo* experiments.** (A) Representative histology examples from each anastomosis tissue operated on with STAR (n = 4), and manual laparoscopic control test (n = 1). The approximate location of each anastomosis is indicated with an arrow. Dashed boxes near each anastomosis represent location of the magnified images. (B) PMN cell as a surrogate measure of inflammation for each sample. (C) Representative examples of the anastomosis collected at necropsy for STAR and control tests.

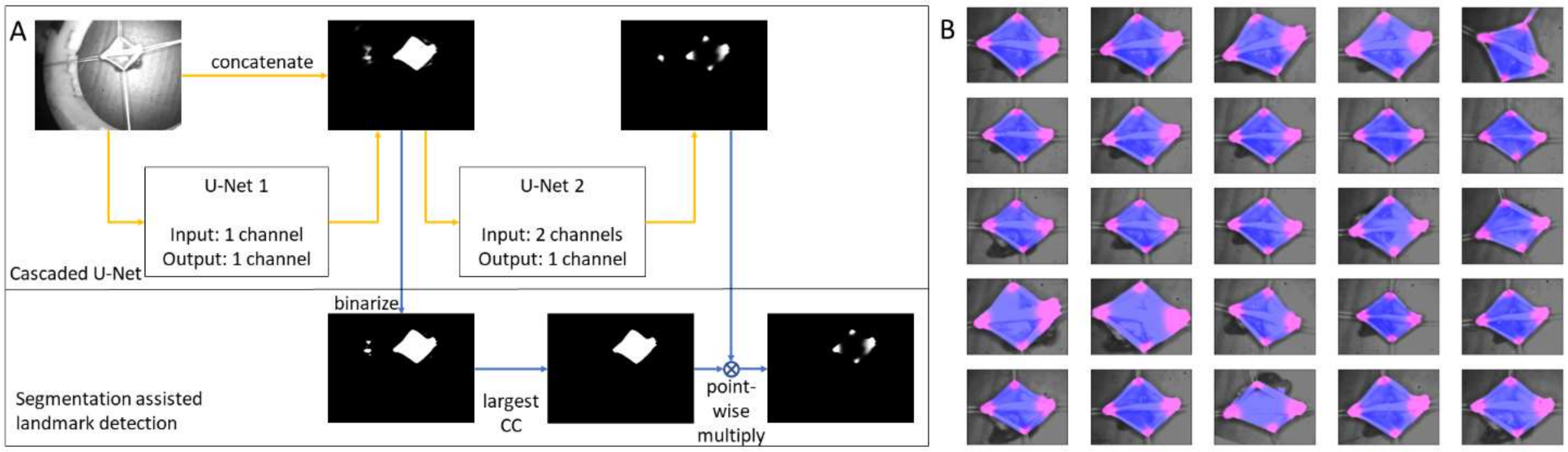

**Fig. 5. Segmentation assisted landmark detection**. (A) Overall data processing scheme based on cascaded U-nets, (B) Final results on the whole testing set. Blue shows the intestine segmentation results, and pink shows the landmark heatmap results.

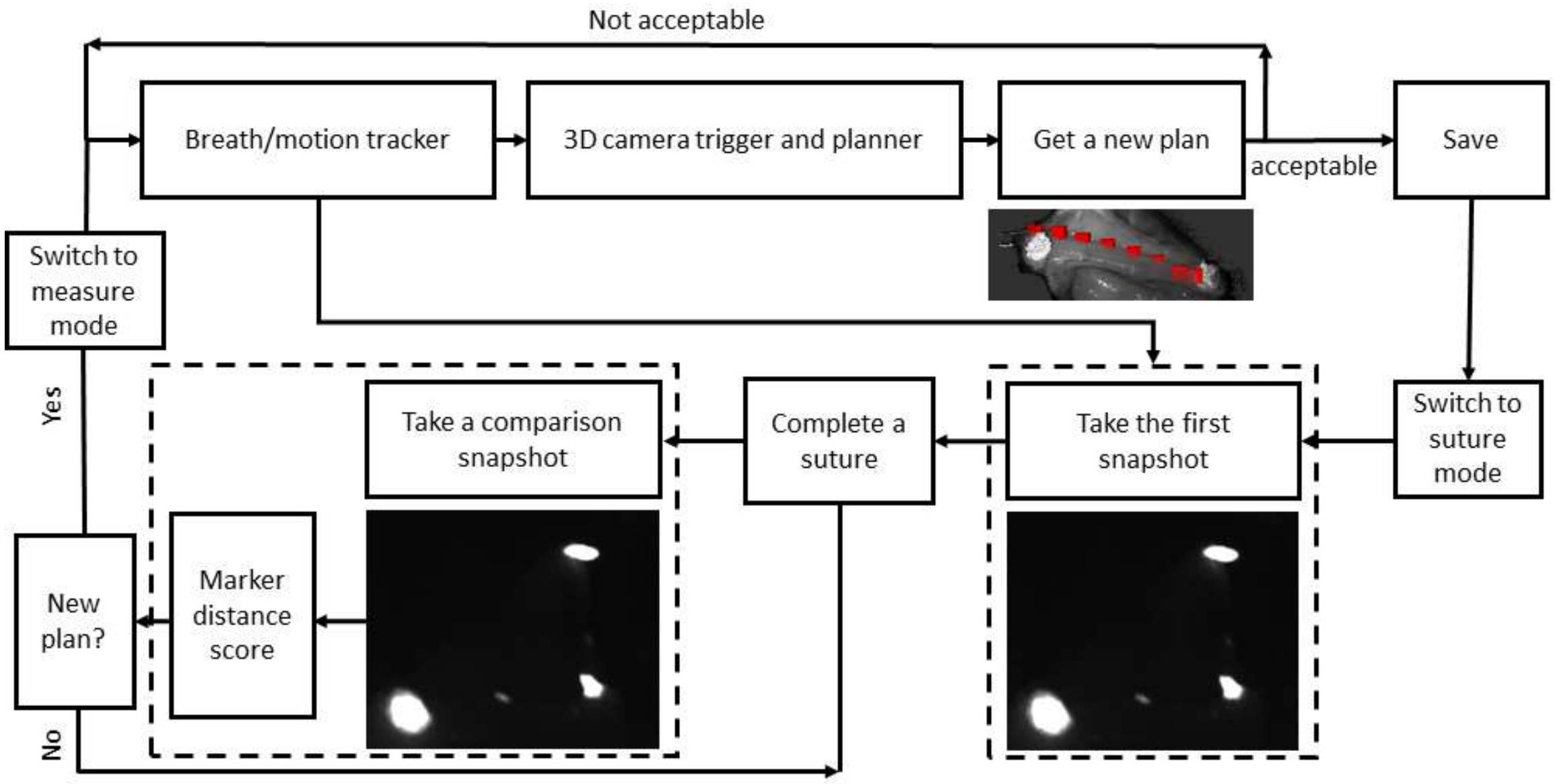

**Fig. 6. Suture (re)planning workflow and deformation tracking.** First, at the stationary point of the breathing cycle, a set of new plans are generated, prefiltered, and presented to the operator. If the plans are approved and selected, robots switch to the suture mode, record the tissue condition before completing a stitch, and compare it to the condition after the stitch. If the tissue deformation score is more than the acceptable threshold, the operator is notified about the need for a new plan and system switches to the measure mode for another round of suture planning. Otherwise, suturing continues with the previous plan.

**Table 1**. ***In vivo* end-to-end anastomosis.** Bowel anastomoses were carried out in pigs using STAR (n = 4) and manual laparoscopic control (n = 1).

| Pig No | Weight at surgery (kg) | Weight at sacrifice (kg) | Leak pressure (psi) | Lumen patency (%) | Completion Time (min) | No of sutures | Suture Hesitancy |
|---|---|---|---|---|---|---|---|
| **STAR 1** | 32.7 | 36.4 | 0.23 | 85 | 59.71 | 24 | 4 |
| **STAR 2** | 35.4 | 37.3 | 0.12 | 85 | 55.64 | 17 | 7 |
| **STAR 3** | 35.3 | 33.5 | 1.2 | 90 | 65.73 | 24 | 11 |
| **STAR 4** | 35.5 | 33.8 | 1.2 | 95 | 67.03 | 21 | 7 |
| **5 (Control)** | 30 | 32 | 1.2 | 90 | 25.6 | 21 | 9 |